# Computer Vision in the Food Industry: Accurate, Real-time, and Automatic Food Recognition with Pretrained MobileNetV2

Shayan Rokhva, Babak Teimourpour*, Amir Hossein Soltani

Shayan_rokhva@modares.ac.ir - Department of Information Technology Engineering, Faculty of Industrial and Systems Engineering, Tarbiat Modares University, Tehran, Iran

B.teimourpour@modares.ac.ir - Department of Information Technology Engineering, Faculty of Industrial and Systems Engineering, Tarbiat Modares University, Tehran, Iran*

Soltani.amirhossein@modares.ac.ir - Department of Information Technology Engineering, Faculty of Industrial and Systems Engineering, Tarbiat Modares University, Tehran, Iran

**ABSTRACT:**

In contemporary society, the application of artificial intelligence for automatic food recognition offers substantial potential for nutrition tracking, reducing food waste, and enhancing productivity in food production and consumption scenarios. Modern technologies such as Computer Vision and Deep Learning are highly beneficial, enabling machines to learn automatically, thereby facilitating automatic visual recognition. Despite some research in this field, the challenge of achieving accurate automatic food recognition quickly remains a significant research gap. Some models have been developed and implemented, but maintaining high performance swiftly, with low computational cost and low access to expensive hardware accelerators, still needs further exploration and research. This study employs the pretrained MobileNetV2 model, which is efficient and fast, for food recognition on the public Food11 dataset, comprising 16643 images. It also utilizes various techniques such as dataset understanding, transfer learning, data augmentation, regularization, dynamic learning rate, hyperparameter tuning, and consideration of images in different sizes to enhance performance and robustness. These techniques aid in choosing appropriate metrics, achieving better performance, avoiding overfitting and accuracy fluctuations, speeding up the model, and increasing the generalization of findings, making the study and its results applicable to practical applications. Despite employing a light model with a simpler structure and fewer trainable parameters compared to some deep and dense models in the deep learning area, it achieved commendable accuracy in a short time. This underscores the potential for practical implementation, which is the main intention of this study.

1. **Introduction**

AI-driven food recognition is crucial in the modern food industry, assisting in tasks like identifying restaurant foods, nutritional tracking, and classifying leftover or wasted food using image data (Lubura *et al.*, 2022). AI visual recognition methods can be used not only for food recognition but also for agricultural applications like identifying and classifying plants, fruits, and vegetables, which are also linked to the food industry (Zhou and Chen, 2023). Therefore, AI applications can significantly boost efficiency or increase processing speed in kitchens and food production facilities (Chakraborty and Aithal, 2024; Lubura *et al.*, 2022; Moumane *et al.*, 2023a). Advanced nations are adopting robotics in kitchens for efficient, minimal-waste operations. Empowered by computer vision and deep learning, these systems recognize food items and enhance productivity in food sectors. Their benefits are amplified by providing accuracy, speed, cost-effectiveness, and scalability for extensive real-world data. (Chakraborty and Aithal, 2024; Fang *et al.*, 2023; Moumane *et al.*, 2023b).

Deep learning, a subset of AI, is superior at solving complex problems, especially with large datasets, which we have in the modern world. Convolutional Neural Networks (CNNs), blending deep learning and computer vision, are widely used in image and video processing. Their performance generally improves with greater datasets. CNNs use convolutional layers to automatically extract and aggregate image features, creating a rich feature map for image classification. They employ max pooling layers for feature selection and non-linear down-sampling, reducing computations. The final classifier determines the data prediction. Compared to CNNs, Artificial Neural Networks (ANNs), with their exponential increase in trainable parameters, require more computational costs, time, and storage, making them less suitable for image classification (Abiyev and Adepoju, 2024; Lubura *et al.*, 2022; Nath and Naskar, 2021).

The past decade has seen the development and application of various CNN models like AlexNet, VGG, GoogleNet, ResNet, EfficientNet, and MobileNet. Each model family differs in characteristics, size, depth, parameters, performance, and computational requirements. This shows that different models can suit different applications (Alzubaidi *et al.*, 2021; Dhillon and Verma, 2020; Liu *et al.*, 2024). Current studies usually focus on enhancing accuracy, reducing computational costs, and tailoring to specific applications (Alzubaidi *et al.*, 2021; Karypidis *et al.*, 2022; Silaparasetty, 2020).

Smart refrigerators, equipped with AI and IoT technologies, can accurately classify food items and proactively mitigate food waste by notifying owners. Despite their current high cost, ongoing research indicates a promising future with more affordable and efficient models, underscoring the potential of image-based food recognition in real-world applications (Gao *et al.*, 2019; Shweta, 2017; Wang *et al.*, 2019). With their internal cameras, they are designed to accurately identify stored items and alert users about the type and freshness of the food. The importance of image-processing models for these appliances has been emphasized in the literature (Gao *et al.*, 2019; Wang *et al.*, 2019). One such study focused on the classification of fruits such as oranges, apples, and bananas, determining their freshness or state of decay (Gao *et al.*, 2019).

Food waste has drawn considerable public and policy attention. Alarmingly, 25-35% of global food production remains uneaten depending on location, leading to resource wastage, greenhouse gas emissions, supply chain disruption, and financial burden on consumers (Ahmadzadeh *et al.*, 2023; de Almeida Oroski and da Silva, 2023; Stenmarck *et al.*, 2016). In this area, image recognition, detection, and segmentation techniques can be highly beneficial. Recognition, the initial and the most influential step is vital for identifying food or food waste types. Its accuracy greatly impacts subsequent processes and any inaccuracies could disrupt the entire operation. Therefore, automatic accurate real-time food recognition and classification should be investigated (Farinella *et al.*, 2020; Lubura *et al.*, 2022; Moumane *et al.*, 2023a).

In this study (Kawano and Yanai, 2014), Deep-CNNs (DCNNs) were used to boost food recognition accuracy by integrating image features, color patches, and Fisher vectors. Another research (Kagaya *et al.*, 2014) addressed the challenges of using classic CNNs and hyper-parameter tuning to enhance food recognition performance. A study utilized GoogleNet, a deep learning model with Inception modules, for food and non-food differentiation (Liu *et al.*, 2016). This study (Zhu *et al.*, 2020) developed a system using a video camera for recognizing stored fruits and vegetables. This model, based on Faster R-CNN, enhanced speed and accuracy over previous methods.

Transfer learning, a key deep learning tool, mitigates data scarcity, cuts computational needs, speeds up model training, and boosts performance. It includes Feature extraction, adjusting only classifier parameters, Fine-tuning, training deeper layers while early ones are fixed, and Full Fine-Tuning, training all parameters and layers. Studying the suitability of models based on the application is crucial. Generally, there are significant differences between deep and dense models compared with lightweight and simpler models in terms of performance, time, computations, and hardware requirements. (Dhillon and Verma, 2020; Farahani *et al.*, 2021; Hosna *et al.*, 2022; Iman *et al.*, 2023).

The study (Siddiqi, 2019) used transfer learning for automated fruit image recognition with VGG16, a deep and dense model, achieving 99.27% accuracy. They (Bansal *et al.*, 2023) improved image classification using VGG19 and transfer learning, achieving 93.73% accuracy on the Caltech-101 dataset with a Random Forest classifier, surpassing previous methods. They (Fakhrou *et al.*, 2021) used DCNNs for smartphone-based food recognition, achieving 95.55% accuracy with transfer learning. Deep ResNets (He *et al.*, 2015) can generally be applied for accurate image classification. For example, to identify crop disease in the agriculture domain, linked to food security, ResNet34 achieved 99.40% accuracy (Kumar *et al.*, 2020). All (Bansal *et al.*, 2023; Fakhrou *et al.*, 2021; Kumar *et al.*, 2020; Siddiqi, 2019) underscored the importance of transfer learning for cutting time and improving performance, specifically using dense models.

Lightweight models also need investigation. In tests with TensorFlow datasets, MobileNetV2 showed superior accuracy over MobileNetV1 (Dong *et al.*, 2020). According to (Dai *et al.*, 2020; Dong *et al.*, 2020; Sahu *et al.*, 2023), MobileNet's efficient architecture and computational performance make it suitable for real-time classification. However, its computational efficiency might slightly compromise accuracy in more complex tasks. Nonetheless, this general statement does not always apply. For instance, using transfer learning and utilizing 108 images sized 256*256, MobileNetV2 was able to accurately classify soil images (Banoth and Murthy, 2024). In another study (Huang and Liao, 2022), LightEfficientNetV2, a lightweight CNN, with fewer parameters, highlighting its speed, achieved 98.33% and 97.48% accuracies on chest X-ray and CT images, respectively. This demonstrated that deep structures, requiring more computation and time, are not always the most accurate.

Quick and accurate image classification can be challenging due to shape, color, specific difficult patterns, and appearance similarities. Yet, CNNs and transfer learning have proven effective (Zhuang *et al.*, 2020). VGG16, MobileNet2, and Resnet50 were tested on a public flower dataset, highlighting the difficulties in classifying similar flowers (Narvekar and Rao, 2020). This is something pivotal to consider as similar challenges regarding color, pattern or shape can occur in food categorization. These challenges should be addressed by further explorations as solutions can positively affect the model's performance (Lubura *et al.*, 2022; Mazloumian *et al.*, 2020).

Several studies were conducted for food waste estimation and computer vision played an important role. In this study (Lubura *et al.*, 2022) a classic VGG-like CNN model was built from the ground up for food recognition, achieving 98.8% accuracy. With image segmentation, Serbia's food waste was found to be 21.3%. In another study, using other image-processing techniques, with transfer learning data augmentation, they classified and segmented food waste with 83.5% and 98.5% accuracy and pixel accuracy, respectively. They also highlighted the importance of accurate and real-time food recognition for food waste reduction (Mazloumian *et al.*, 2020). Various models were used for waste detection, segmentation, and volume estimation, also considering organic waste such as food waste. Simpler and lightweight models were also found to be beneficial (Geetha *et al.*, 2022).

Despite advancements in automatic food recognition, the challenge remains to achieve fast, accurate results with minimal computational cost and less dependency on expensive hardware accelerators, particularly for efficient deployment in real-world embedded and portable devices (Ahmadzadeh *et al.*, 2023; Dong *et al.*, 2020). This research aims to enhance accurate and real-time food identification using an optimized lightweight model like MobileNetV2. Techniques such as transfer learning, data augmentation, regularization, dynamic learning rate, hyperparameter tuning, and varying image sizes are also used to enhance the model's robustness and performance, in turn filling a gap and contributing to real-world applications.

This research's key contributions include:

1. Efficient application of MobileNetV2, a CNN model known for low computational demands and faster processing.
2. Utilizing the public Food11 dataset from Kaggle, with 16643 images across 11 food categories, highlighting inter-category similarities and intra-category diversities.
3. Implementing transfer learning in full fine-tuning mode to enhance performance, accelerate the model, save time, and address data scarcity.
4. Application of data augmentation, regularization, and hyperparameter tuning to prevent overfitting, stabilize accuracy, and improve real-world applicability.
5. Consideration of different food image sizes to balance metric satisfaction and speed.
6. Detailed discussion offering valuable insights into the findings and suggestions for further improvements.

The rest of the study is organized as follows: Section 2 explains the methodology, Section 3 provides results and analysis, Section 4 discusses findings deeply, and Section 5 concludes the paper.

## 2. Materials and Methods

### 2.1. Schematic Diagram of the Research Workflow

Figure 1 offers a general visual summary of our research methodology, delineating the key stages of the applied approach.

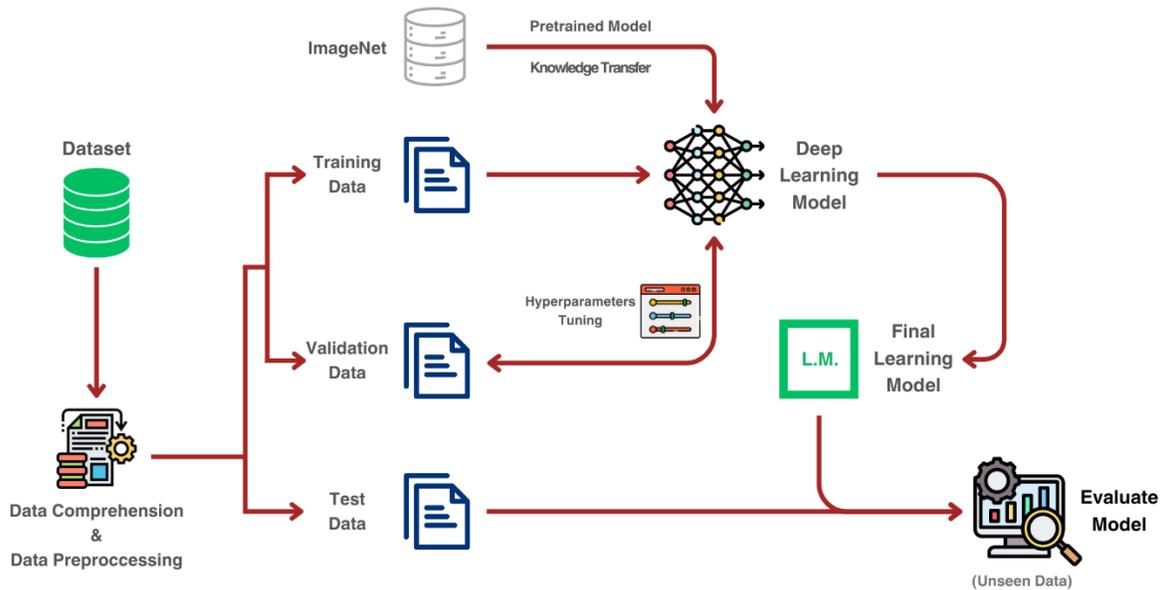

Figure 1- General visual summary of the research methodology

**2.2. Hardware Configuration**

This research utilized Google Colab's T4 GPU, boasting 16GB memory and up to 260 TOPS of computational power.

**2.3. Data Acquisition, Distribution & Comprehension**

Dataset comprehension is influential for effective research and analysis (Alvi, 2023; Wang and Zhu, 2023). The Food11 dataset, sourced from Kaggle ("Food-11", n.d.), had originally been divided into training, validation, and evaluation subsets, preventing data leakage (Brookshire *et al.*, 2024). These subsets contain 9866, 3439, and 3347 images across 11 categories.

Table 1 breaks down the Food 11 dataset. It shows the number of images per food category and subset, such as 994 and 362 "Bread" images in training and validation, respectively. It also presents the distribution of each food category across subsets (e.g., "Dairy Product" is 59.6% in training, 20% in validation, and 20.5% in evaluation) and different food categories within a specific subset (e.g., "Bread" is about 10% of all training data).

Table 1 indicates a 60% allocation of all food category samples to the training subset, with the remainder divided almost equally between validation and evaluation. However, an imbalanced distribution is observed within each subset and the dataset. For example, "Rice" accounts for less than 3% of images in all subsets, while "Meat" and "Dessert" make up about 13% and 15% respectively. This imbalance, particularly in the training subset, could affect the model's learning for certain categories. Therefore, further exploration is required via a confusion matrix analysis.

Table 1 – Data distribution of the Food 11 dataset

| | Training | Validation | Evaluation | Total / Sum |
|---|---|---|---|---|
| **Bread** | 994 | 362 | 368 | 1724 |
| **(in each class)** | (57.6%) | (21%) | (21.4%) | (100%) |
| **(in each subset)** | (10%) | (10.5%) | (11%) | (10.4%) |
| **Dairy Product** | 429 | 144 | 148 | 721 |
| **(in each class)** | (59.5%) | (20%) | (20.5%) | (100%) |
| **(in each subset)** | (4.3%) | (4.2%) | (4.4%) | (4.3%) |
| **Dessert** | 1500 | 500 | 500 | 2500 |
| **(in each class)** | (60%) | (20%) | (20%) | (100%) |
| **(in each subset)** | (15.2%) | (14.6%) | (14.9%) | (15%) |
| **Egg** | 986 | 327 | 335 | 1648 |
| **(in each class)** | (59.8%) | (19.9%) | (20.3%) | (100%) |
| **(in each subset)** | (10%) | (9.5%) | (10%) | (9.9%) |
| **Fried Food** | 848 | 326 | 287 | 1461 |
| **(in each class)** | (58%) | (22.3%) | (19.7%) | (100%) |
| **(in each subset)** | (8.6%) | (9.5%) | (8.6%) | (8.8%) |
| **Meat** | 1325 | 449 | 432 | 2206 |
| **(in each class)** | (60.1%) | (20.3%) | (19.6%) | (100%) |
| **(in each subset)** | (13.4%) | (13.1%) | (12.9%) | (13.2%) |
| **Noodles-Pasta** | 440 | 147 | 147 | 734 |
| **(in each class)** | (60%) | (20%) | (20%) | (100%) |
| **(in each subset)** | (4.6%) | (4.3%) | (4.4%) | (4.4%) |
| **Rice** | 280 | 96 | 96 | 472 |
| **(in each class)** | (59.4%) | (20.3%) | (20.3%) | (100%) |
| **(in each subset)** | (2.8%) | (2.8%) | (2.9%) | (2.8%) |
| **Seafood** | 855 | 347 | 303 | 1505 |
| **(in each class)** | (56.8%) | (23.1%) | (20.1%) | (100%) |
| **(in each subset)** | (8.7%) | (10.1%) | (9.1%) | (9%) |
| **Soup** | 1500 | 500 | 500 | 2500 |
| **(in each class)** | (60%) | (20%) | (20%) | (100%) |
| **(in each subset)** | (15.2%) | (14.6%) | (14.9%) | (15%) |
| **Vegetable-fruit** | 709 | 232 | 231 | 1172 |
| **(in each class)** | (60.5%) | (19.8%) | (19.7%) | (100%) |
| **(in each subset)** | (7.2%) | (6.8%) | (6.9%) | (7.1%) |
| **Total / Sum** | 9866 | 3430 | 3347 | 16643 |
| **(in each class)** | (59.3%) | (20.6%) | (20.1%) | (100%) |
| **(in each subset)** | (100%) | (100%) | (100%) | (~100%) |

### 2.4. Data Transformation & Visualization

This study used two types of data transformations: Main and Augmenting. The Main transformation, including Resizing, Tensoring, and Normalization, was applied to validation and evaluation subsets. The Augmenting transformation, which includes main transformation with additional techniques like Random Rotation, Horizontal Flipping, Color Jittering, and Erasing, was used during training to enhance data diversity, not quantity, to improve model performance, and robustness, and to reduce overfitting. We intentionally decided not to increase the quantity of data and not to balance it to evaluate its effects and discuss it. Yet, data diversity was improved.

During Normalization, transfer learning allowed the use of ImageNet's mean and Standard deviation (STD). The technique did not exactly normalize the training set's mean and STD to 0 and 1 but approximated them closely, suitable for Food11 dataset images, which are in red, green, and blue channels (RGB), and similar to ImageNet data (Huh *et al.*, 2016).

The study also explored how image resolution (32, 64, 128, 256) affects model speed and performance. As per Figure 2, larger images enhanced clarity, but the clarity difference between two consecutive resolutions reduced as image size grew. High-resolution images (512 and 1024) were avoided due to their adverse impacts on model speed (Thambawita *et al.*, 2021), particularly for real-time and non-sensitive applications such as food recognition.

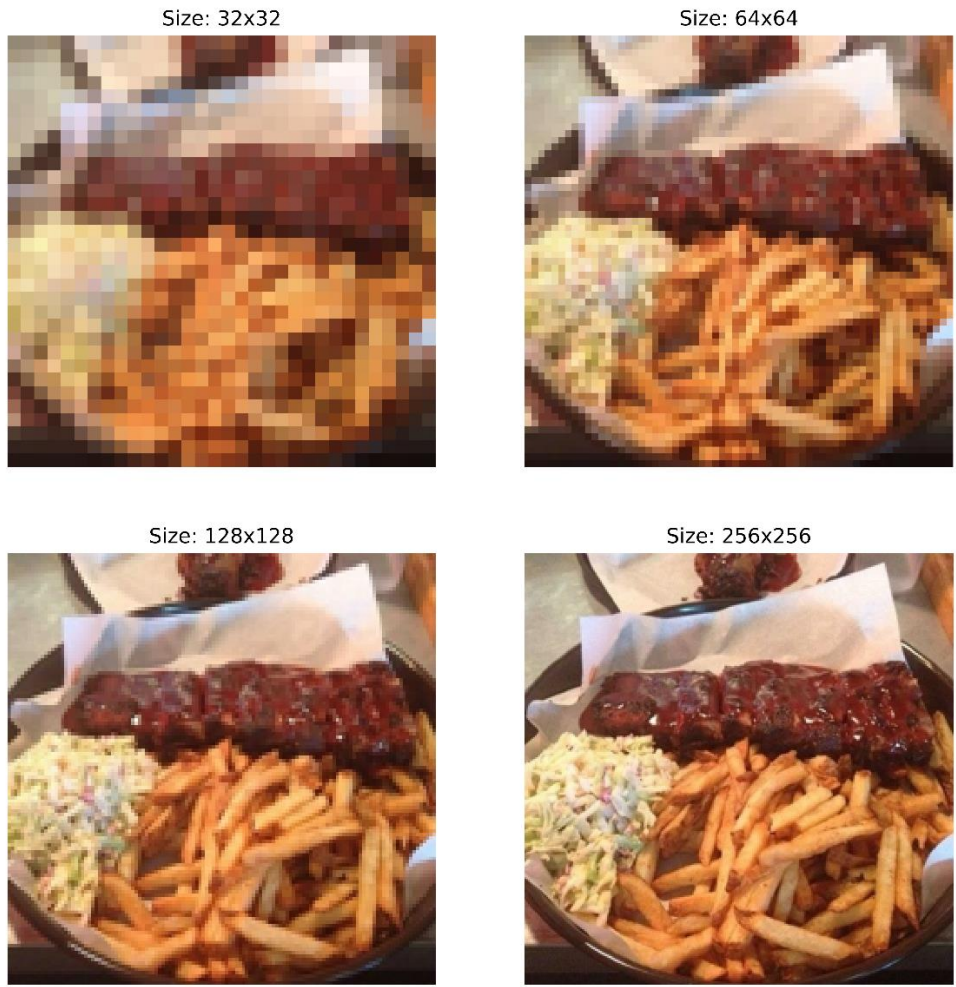

Figure 2 – Image representation across multiple resolutions

### 2.5. Model

MobileNetV2 was employed due to its efficiency and speed caused by a combination of depth and point-wise convolution, linear bottlenecks, and shortcut connections. In many applications, it could outperform MobileNetV1 while having fewer parameters and computations. Furthermore, it is more popular than other MobileNet models. (Ai *et al.*, 2021; "MobileNetV2", 2018; Sandler *et al.*, 2018; Wu *et al.*, 2023).

Figure 3 shows MobileNetV2 has the fewest parameters (~3.5 million) among the mentioned ImageNet classification models, making it ideal for our study. Our customized version for the 11-class problem had 2.238 million parameters due to 11-class classifier modification, which made it even more lightweight.

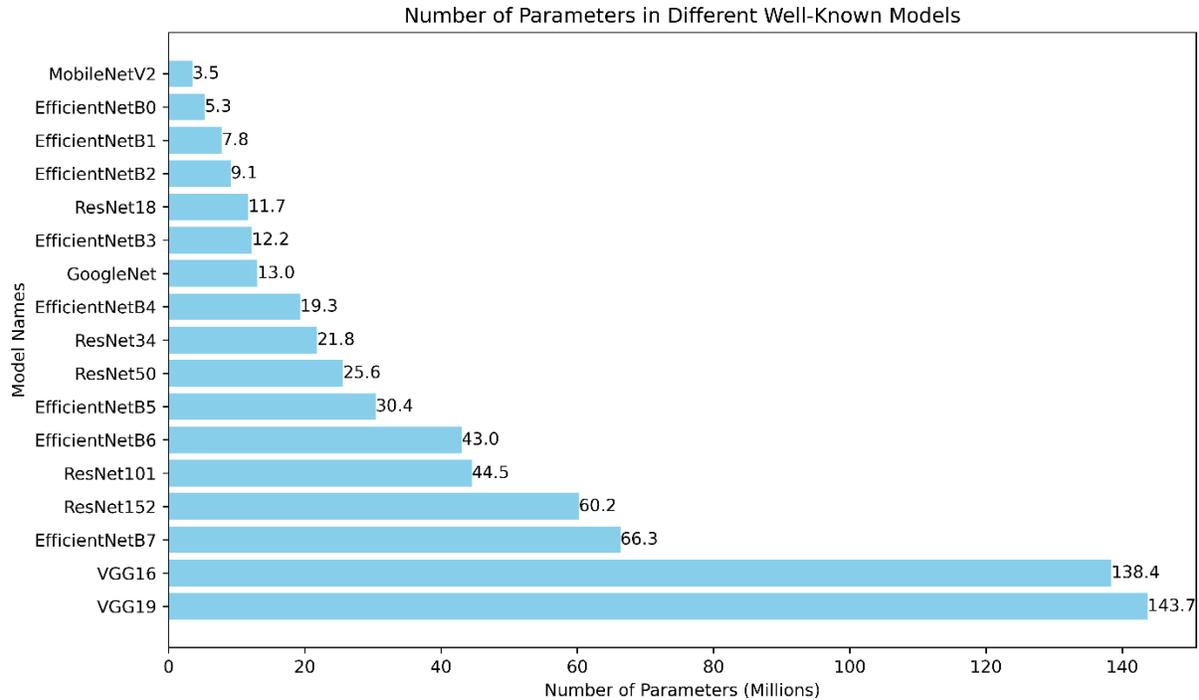

Figure 3 – Approximate parameter counts for some well-known models based on ImageNet

### 2.6. Transfer Learning

Transfer learning was used to enhance model performance and speed, conserve resources, and even address probable data scarcity. It is most effective when source and target datasets have similar features, and with enough data in the understudying dataset, full fine-tuning usually yields superior performance as it is more customized (Hosna *et al.*, 2022; Zhuang *et al.*, 2020). Therefore, we used a pre-trained MobileNetV2 model, customized it for an 11-class task, and benefited from full fine-tuning across 30 epochs. Given the similarities between Food11 and ImageNet, and that both datasets are RGB, effective transfer learning was expected.

### 2.7. Loss Function, Optimizer, and Hyperparameter Tuning

In this multiclass classification task, Cross Entropy Loss was used as the loss function and Stochastic Gradient Descent (SGD) as the optimizer. Several hyperparameters were tuned, including learning rate (LR), momentum, L2 regularization, and Nesterov accelerator for optimal results.

Momentum was set at 0.9 to help avoid local minima and achieve minimized loss and optimal performance. The Nesterov accelerator was enabled for better convergence. Learning rate, the most influential hyperparameter, was carefully chosen to ensure smooth convergence, avoiding divergence, and unnecessary fluctuations. Initially, LR=0.001 was found to provide better and faster convergence. L2 regularization, important for reducing overfitting and improving generalization, was also tuned. To reach optimal performance a grid of different learning rates and L2 regularization values was searched and LR=0.001 and WD=0.0001 were found to be suitable for all image resolutions in this study.

To even further optimize, a learning rate scheduler was used to dynamically adjust the learning rate. Through experiments, we found it beneficial to reduce the learning rate by a factor of ten every 10 epochs. The Training, Validation, and Evaluation batch sizes that were used were 64, 128, and 128 respectively.

### 2.8. Robust Generalization & Alternative for K-fold Cross Validation

Addressing real-world problems requires generalization and applicability. To ensure generalizable results with the Food11 dataset, which had been inherently pre-divided into training, validation, and testing subsets and did not permit K-fold cross-validation, a method inspired by 5-fold shuffled cross-validation was used. The code was executed five times from scratch for each image size and the average performance was reported. Each execution involved a fresh initialization of all parameters, the loss function, the model, and the optimizer, preventing data leakage. This approach improves the study's generalizability by addressing randomness in data transformation and data loader creation. It should be noted that the results, derived from executing the code five times, were anticipated to be similar for each image size.

### 2.9. Employed Evaluation Metrics

The utilized Food11 dataset was moderately, not extremely, imbalanced. We found accuracy sufficient for our needs, as other metrics like Precision, Recall, and F1 score are more relevant in cases of severe imbalance or when certain misclassifications have severe consequences. However, a normalized confusion matrix was also used to identify common misclassifications and to provide a comprehensive and understandable evaluation of misclassifications, which can be discussed further.

### 3. Results & Analysis

Table 2 succinctly encapsulates the study's outcomes, emphasizing the evaluation subset's accuracy and speed across varying image dimensions. For brevity and clarity, we confined our report to the model's performance metrics—accuracy and speed—solely on the evaluation subset, representing unseen data that the model comes across in the real world. Each image size underwent five experiments, documenting the evaluation's accuracy and speed. The mean values of these metrics for specific image resolutions were computed to provide average speed and accuracy. Furthermore, for deeper comprehension, Figure 4 depicts the accuracy progression trend for various image sizes across 30 epochs, offering insights into model fitting, smooth convergence, optimal performance, employed techniques' impacts, and the influence of image resolution.

Table 2 – Findings in terms of accuracy and speed for evaluation subset

|  | Evaluation accuracies for 5 experiments (percent) | Evaluation speeds for 5 experiments (Image/s) | Average evaluation accuracy (percent) | Average evaluation speed (Image/s) |
| --- | --- | --- | --- | --- |
| Image size = 32 | 59.81 – 61.06 – 59.71 – 60.11 – 60.18 | 683.5 – 698.9 – 697.6 – 725.7 – 706.5 | **60.17%** | **702.4** |

| Image size = 64 | 77.86 – 77.69 – 77.96 – 76.89 – 76.91 | 611.8 – 636.2 – 637.7 – 627.2 – 620.8 | **77.46%** | **626.7** |
| --- | --- | --- | --- | --- |
| Image size = 128 | 87.78 – 88.41 – 88.12 – 88.68 – 88.02 | 451.8 – 445.5 – 454.4 – 450.9 – 458.2 | **88.20%** | **452.1** |
| Image size = 256 | 92.98 – 92.81 – 93.05 – 92.71 – 93.29 | 293.1 – 293.1 – 289.3 – 291.9 – 288.1 | **92.97%** | **291.1** |

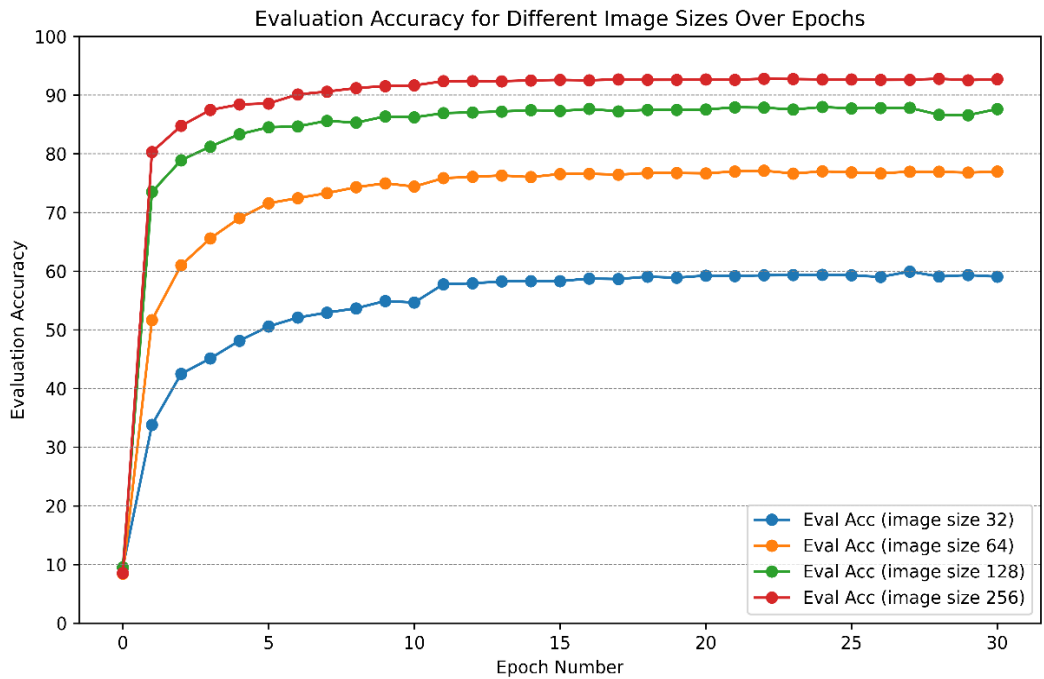

Figure 4 – Evaluation accuracy progression trend across 30 epochs for different image resolutions

Table 2 reveals that larger image sizes improved accuracy but decreased speed, increasing time. Notably, utilizing 256*256 images the simple MobileNetV2 model could classify 291 images with ~92.97% accuracy in a second, balancing speed and accuracy effectively. Additionally, both Table 2 and Figure 4 show that accuracy gains, the amount of rise caused by using higher image sizes, decreased when image quality increased. Hence, a much smaller improvement in accuracy with 512*512 images compared to the ~5% improvement from 128*128 to 256*256 images is expected. Table 2 also indicates that while for higher image resolutions accuracy gain between two consecutive image resolutions gets smaller, speed reduction gets noticeably greater. Thus, extremely high-quality images should be avoided.

Figure 5 presents the confusion matrix for the best image size of 256*256. It must be stated that this matrix was generated from a single code execution, and due to the inherent randomness, minor variations might occur in the matrices and internal accuracies. For example, some classes' accuracies may fluctuate up or down to a maximum of 2%. However, these variations were found to be nearly negligible for evaluating and interpreting the results, specifically for higher image sizes. This ensures the validity and applicability of the results and subsequent discussions. Table 2 also confirms this point, showing that the accuracy disparity, the difference between minimum and maximum values for each image size, for image sizes 32, 64, 128, and 256 was 1.35%, 1.07%, 0.9%, and 0.58% respectively, indicating that higher image qualities also reduced the effect of randomness.

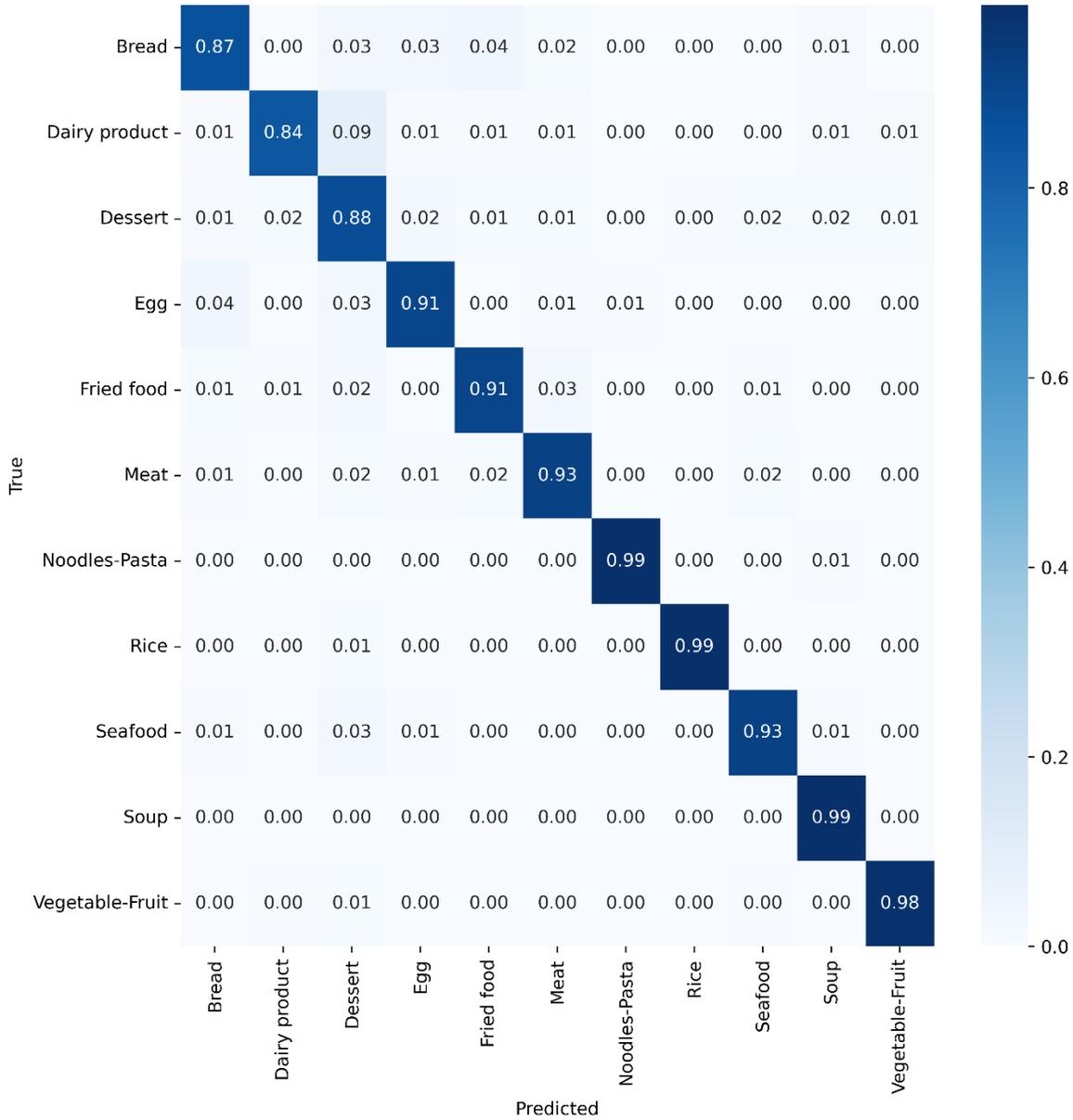

Figure 5 – Confusion matrix for 256*256 image resolution

While Table 2 shows that the model prediction for 256*256 images was approximately 93% accurate, Figure 5 reveals where misclassifications lay, providing more details. Initially, Figure 5 illustrates that predictions were highly satisfactory, impressively 8 out of 11 classes surpassed 90% accuracy and 10 out of 11 surpassed 85% accuracy. Moreover, it reveals that some classes, such as "Soup", "Rice", and "Noodles-Pasta", were significantly better predicted than some other classes like "Dairy-Products", "Bread", and "Dessert". It seems as if some classes were easier to predict while others were more challenging. Additionally, underperformed classes did not only show more misclassifications but also showed most misclassifications among themselves. For instance, "Dairy-Products" were more likely to be wrongly predicted as "Dessert". Since these points of contention can impact model performance, they should be discussed in depth.

## 4. Discussion

### 4.1. Studying the Effectiveness of Employed Techniques

#### 4.1.1. Transfer Learning

The influence of transfer learning first became apparent during hyperparameter tuning. An initial learning rate of 0.001 was deemed suitable for the early epochs. This small value was a result of utilizing the pre-trained model. Without transfer learning, the anticipated learning rate would have been greater, possibly tenfold. The second impact is noticeable in Figure 4. At Epoch=0, all weights were random, leading to an accuracy of around 9% for all image qualities, as expected for an 11-class classification (100/11). However, after just one epoch of pre-trained model and full fine-tuning, the accuracies dramatically increased, hitting almost 80% for images of 256*256 resolution. This demonstrates the efficacy of transfer learning in achieving high, and sometimes sufficient, accuracies, while eliminating the need for additional time and costly hardware. Furthermore, Figure 4 illustrates that while transfer learning was advantageous for all image qualities, its effects were particularly pronounced on high-quality images.

#### 4.1.2. Data Augmentation and Regularization

Data Augmentation and L2 Regularization were used to prevent overfitting, improving model robustness and generalization. This allowed extended training without early stopping concerns. Figure 4 confirms their effectiveness as evaluation accuracies showed no overfitting, allowing the model to train until reaching the optimal point.

#### 4.1.3. Dynamic Learning Rate

The initial learning rate was found 0.001 for optimal performance. All accuracies saturated after 10 epochs (Figure 4), showing the need for a dynamic LR. Therefore, LR was reduced tenfold after 10 epochs. Thus, epochs 11-20 and 21-30 had an LR of 1e-4 and 1e-5 respectively. This increased the previously-saturated accuracy for all sizes at epoch 11. However, the gain was much smaller between epochs 20 and 21 due to the nature of small LR at 1e-5.

### 4.2. Studying Image Quality

Table 2 shows that as image quality improved, accuracy rose but speed dropped. However, higher-quality images yielded less accuracy gain but significantly reduced speed, emphasizing the need for optimal image quality selection. In this study, 256 was found to be the ideal balance between high accuracy and brilliant speed.

Table 2 shows that the effect of randomness diminished by employing good-quality images, as observed by the differences in accuracies among various tries for a specific image size. This is because by utilizing good-quality images CNN can extract better and more meaningful features, resulting in better performance, also being less affected by randomness. This is more evident for MobileNetV2 with a less dense algorithm and simpler convolution.

Figure 4 indicates that transfer learning led to better performance and faster convergence for higher-quality images. This is probably because pre-trained MobileNetV2 has been trained on ImageNet's good-quality images, providing better weights for higher image qualities.

### 4.3. Studying Confusion Matrix and Dataset Challenges

The Food 11 dataset, utilized for this study, poses a significant challenge due to the intra-category diversity and inter-category similarities. For instance, the "Fried Food" and "Meat" categories overlap in images like Figure 2, which contains both french fries and meat but is categorized as "Meat". A small salad presence could also be biased toward the "Vegetable-Fruit" category. Furthermore, the "Bread" category contains numerous pizzas and normal pieces of bread simultaneously. These examples are very frequent, representing complexities in the dataset, making prediction challenging.

As delineated in Section 4, Figure 5 reveals that certain classes, such as "Soup" and "Vegetable-Fruits", "Rice", and "Noodles-Pasta" were classified accurately. Conversely, classes like "Bread", "Dairy Products", and "Dessert" exhibited lower performance relatively. Notably, "Dairy Products" were frequently misclassified as "Desserts", which adversely impacted the overall accuracy. Despite the common perception of neural networks as 'black boxes', insightful discussions and recommendations can indeed be beneficial.

Data imbalance and scarcity are critical factors to consider, significantly affecting the model's learning process and class performance. More training data of a specific class generally improves the feature extraction and learning process, increasing the accuracy of that class, while less data does the opposite in general. However, the observed misclassification pattern did not appear to be directly attributable to this phenomenon. Table 3 shows that "Noodles-Paste" and "Rice", despite their small contributions, performed excellently, matching "Soup", which has the largest proportion. Moreover, although "Soup" and "Desserts" contributed evenly by 15%, they performed completely differently. Also, "Noodels-Pasta" and "Dairy-Products" contributed almost evenly, but had the most and least performance respectively. There are also more examples to consider. Hence, at least in this study, prediction accuracies for a class do not heavily rely on the data volume phenomenon.

Table 3 – The proportion of categories in the dataset and their corresponding accuracy

| Food Category / Classes | The proportion in the dataset (~portion in each subset) (Table 1) | The accuracy achieved on 256-sized images (Figure 5) |
|---|---|---|
| Bread | 10.4% | 87% |
| Dairy Products | 4.3% | 84% |
| Dessert | 15% | 88% |
| Egg | 9.9% | 91% |
| Fried Food | 8.8% | 91% |
| Meat | 13.2% | 93% |
| Noodles-Pasta | 4.4% | 99% |
| Rice | 2.8% | 99% |
| Seafood | 9% | 93% |
| Soup | 15% | 99% |
| Vegetable-Fruit | 7.1% | 98% |

Our hypothesis attributes the observed differences to the food categories' color patterns and overall visual characteristics. For example, the watery nature of "Soup" simplifies its classification. "Vegetables-Fruits" often exhibit green hues or specific colors common to most fruits. Similarly, the predominance of brown, black, and red in "Meat" facilitates its classification. "Rice" is grainy and comes mostly in white or brown colors, which facilitates its classification. Also, "Noodels-Pasta" can be assumed as a filamentary food category. Therefore, these categories tend to perform well.

Conversely, food categories such as "Bread", "Dairy Products", "Desserts", "Egg", and "Fried Food" typically exhibit bright color patches of yellow, orange, light brown, and white. These similar colors and patterns also exacerbate inter-category similarities. If our hypothesis holds, these classes are not only anticipated to yield lower performance but also are prone to mutual misclassification. This is particularly more relevant when employing MobileNetV2, which utilizes simpler convolutions compared to more complex models like ResNet50 or EfficientNetB7.

Table 7's outcomes support our conjectures, but it is crucial to remember that these are hypotheses, backed by dataset understanding and observations. While the hypothesis aligns with the results, it may not be the sole determinant.

### 4.4. Comparison with Similar Studies

This section delves into research endeavors that parallel our objective of accurate Food-11 image classification. The investigation (Özsert Yiğit and Özyildirim, 2018) utilized CaffeNet and AlexNet for food image classification. Using Stochastic Gradient Descent (SGD) as an optimizer, they attained accuracies of 80.51% and 82.07% respectively. Interestingly, when they switched to Adam as the optimizer, the same models demonstrated improved accuracies of 83.7% and 86.92% respectively. The image resolution used in their study was 512*512.

In a study (Islam *et al.*, 2018a), the researchers utilized a pre-trained InceptionV3 model coupled with an effective data transformation technique for food image classification of the Food-11 dataset. The images used in this study had dimensions of 299*299 after pre-processing. The researchers were successful in achieving an accuracy of 92.86%. In a subsequent research (Islam *et al.*, 2018b), a CNN was built from scratch. Pretrained InceptionV3 was also used as a part of the study and the image dimensions in this study were 224*224. After applying various techniques, the proposed method yielded an accuracy of 74.7%.

In this research (Suddul and Seguin, 2023), the initial approach was to construct a CNN from the ground up. However, this approach yielded a subpar accuracy of 64.6%. To enhance their results, they incorporated a pre-trained EfficientNetB2 model, leveraging the benefits of transfer learning. Additionally, they employed data augmentation strategies to manage the imbalanced dataset and prevent early stopping. Ultimately, they achieved 94.5% accuracy and 94.7% F1 score.

In the recent research (Bu *et al.*, 2024), combining data transformation, transfer learning, and ensemble learning models: VGG19, ResNet50, MobileNetV2, and AlexNet, all pre-trained on ImageNet, the researchers could reach a perfect accuracy of 96.88%, surpassing all prior studies on the Food-11 dataset.

Leveraging a range of prestigious deep learning techniques and considering multiple image resolutions, our focus was on achieving harmony between performance and speed. This approach enabled us to attain an impressive ~92.97% accuracy for real-time analysis in practical applications using 256-sized images. Not only performance but also time and speed were key considerations in our approach. Approximately 88.2% accuracy was also achieved using 128-sized images. By employing a streamlined lightweight model like MobileNetV2, which has fewer than 2.4 million parameters in this task, we were able to achieve remarkable results with brilliant time and speed. We also investigated the effectiveness of employed techniques, remarkable misclassifications, and their probable reasons.

### 4.5. Policy Implications

Given the impressive speed and high accuracy demonstrated, the study's results could pave the way for the development of affordable devices for food recognition in food production and consumption scenarios. This has significant implications for tracking nutrition, food, and food waste. It can be extremely advantageous for AI-operated kitchens that prioritize swift processes and minimal food waste, a feature sought after by many developed countries (Chakraborty and Aithal, 2024). As suggested (Gao *et al.*, 2019; Zhu *et al.*, 2020), cameras in smart refrigerators can play a pivotal role in food waste reduction and sustainability, and this research suggests that pre-trained MobileNetV2 can also be a good option. Lastly, objectives and potential real-world applications encapsulate the essence of our study.

### 4.6. Suggestions for Future Work and Further Improvement

- ➢ More future research can concentrate on accurate and real-time food recognition, utilizing or developing lightweight models or even deep dense models like EfficientNets or ResNet101 with suitable transfer learning methods.
- ➢ Investigating misclassified classes due to factors like similar patterns, colors, and data scarcity is pivotal. Research should also focus on understanding and addressing inter-category similarities and intra-category diversities.
- ➢ Ensemble learning, known for its improvement, could be valuable. Using multiple lightweight models together may enhance performance without increasing time.
- ➢ Applying knowledge distillation to lightweight models can improve performance without significantly slowing down the speed, making it a compelling research area in image classification.

➢ This research centered on automatic food recognition, a classification task. Other computer vision methods like detection and segmentation could be employed to tackle global issues like food waste.

5. **Conclusion**

This research leveraged a pre-trained MobileNetV2 model, complemented by full fine-tuning, data augmentation, hyperparameter optimization, dynamic learning rate, and the use of images of varying quality for automatic, accurate, and real-time food recognition. The adopted approach proved effective in mitigating overfitting, reducing performance variability, and enhancing the model's overall performance and generalizability. The findings revealed that using images of size 256, the model could classify numerous images per second with an accuracy rate of approximately 93%. According to the results, using images larger than this size will not significantly improve accuracy as accuracy gains diminish for very high image qualities, yet speed drops significantly. The study underscores the efficacy of cost-effective pre-trained MobileNetV2, coupled with appropriate techniques, in delivering high speed and accuracy. Despite the dataset's inherent challenges, such as inter-category similarities and intra-category diversities, which were discussed in detail, the results are commendable and applicable to real-world, real-time applications.

**Employing Color for Hardcopy Printing**

We suggest that Figures 2, 4, and 5 should be presented in color for paper printing..

**Decalaration of Interest**

None.

**Declaration of Generative AI and AI-assisted technologies in the writing process**

During the preparation of this work, the authors used "Microsoft Copilot" solely in order to enhance language clarity and readability. After using this tool/service, the authors reviewed and edited the content as needed and take full responsibility for the content of the publication.

**Funding**

This research is not funded by anyone.